\documentclass[twocolumn, 8pt]{article}
\usepackage{authblk}
\usepackage{JAMIA}
\usepackage{amsmath}
\usepackage{textcomp} 
\providecommand{\textsubscript}[1]{$_{\textrm{#1}}$}
\title{Planner--Auditor Twin: Agentic Discharge Planning with FHIR-Based LLM Planning, Guideline Recall, Optional Caching and Self-Improvement}
\author[1]{Kaiyuan Wu\textsuperscript{*}}
\author[2]{Aditya Nagori, PhD}
\author[1,2,3,4]{Rishikesan Kamaleswaran, PhD\textsuperscript{*}}

\affil[1]{Department of Computational Biology and Bioinformatics, School of Medicine, Duke University, Durham, NC, 27710, USA}
\affil[2]{Department of Surgery, School of Medicine, Duke University, Durham, NC, 27710, USA}
\affil[3]{Department of Electrical and Computer Engineering, School of Medicine, Duke University, Durham, NC, 27710, USA}
\affil[4]{Department of Biomedical Engineering, School of Medicine, Duke University, Durham, NC, 27710, USA}

\affil[*]{\vspace{0.5em} \newline \textbf{*Corresponding Authors:}\
Kaiyuan Wu (\texttt{kw459@duke.edu})\\
Rishikesan Kamaleswaran, PhD (\texttt{r.kamaleswaran@duke.edu})}
\date{}

\begin{document}

\par\noindent\rule[-7pt]{15.5cm}{0.2em}
\begin{strip}
    \begin{minipage}{.88\textwidth}
        \maketitle
        \vspace{0.1em}
        \small
        \abstractSection
        {Large Language Models (LLMs) show great potential for clinical discharge planning but are limited by risks of hallucination and miscalibration. This study introduces the self-improving, cache-optional "Planner--Auditor" framework that improves safety and reliability by decoupling generation from deterministic validation and targeted replay.} 
        {We implemented an agentic, retrospective, FHIR-native evaluation pipeline using the MIMIC-IV-on-FHIR dataset. For each patient, the Planner (LLM) generates a structured discharge action plan with an associated confidence estimate. The Auditor is a deterministic module that scores multi-task coverage, tracks calibration (e.g., Brier/ECE proxies), and monitors action-distribution drift. Two-tier self-improvement is supported: (i) within-episode regeneration when enable\_self\_improve is enabled (logic handled inside the planner pipeline), and (ii) cross-episode discrepancy buffering and replay for high-confidence, low-coverage cases.} 
        {While context caching yielded notable improvements over the baseline, the self-improvement loop was the primary driver of performance, significantly increasing task coverage from 32\% to 86\%. Calibration metrics (Brier score and ECE) were markedly reduced, and the incidence of high-confidence misses substantially lowered. The discrepancy buffer further repaired stubborn high-confidence omissions during replay.} 
        {The findings indicate that feedback-driven regeneration and targeted replay are effective control mechanisms for reducing omissions and improving confidence reliability in structured clinical planning. Separating an LLM Planner from an observational, rule-based Auditor enables systematic reliability measurement and safer iteration without model retraining.} 
        {The Planner--Auditor framework offers a practical path toward safer automated discharge planning using interoperable FHIR data access and deterministic auditing, supported by reproducible ablations and reliability-focused evaluation.} 
        {Large Language Models, Clinical Discharge Planning, HL7 FHIR, Patient Safety, Confidence Calibration, Agentic Workflows, Self-Improvement} 

        \par\noindent\rule[-7pt]{15.5cm}{0.2em}
        \hspace{2cm}
    \end{minipage}
\end{strip}

\section*{1. Background and Significance}

The administrative burden of clinical documentation is a primary cause of clinician burnout\cite{ref1,ref2}, with hospital discharge planning representing a complex and high-stake task\cite{ref3,ref4}. This process involves coordinating multidisciplinary teams, scheduling follow-up services, reconciling medications, and educating patients -- all under time pressure as patients transition from hospital to home\cite{ref5,ref6,ref7}. Even with decades of improvement, hospitals worldwide still struggle with non-medical discharge delays\cite{ref4,ref8}, fragmented coordination\cite{ref9}, and unpredictable lengths of stay\cite{ref10,ref11,ref12}. Communication gaps at discharge can directly impact patient safety, while inadequate instructions and preparations are associated with medication errors and readmissions after discharge\cite{ref13}. For instance, over half of patients misunderstand some aspect of their post-discharge medications\cite{ref14,ref15}, and nearly one in five patients experience an adverse event within weeks of discharge, many of which are preventable with better planning\cite{ref16,ref17}. These challenges necessitates the need for advanced tools to support safer, more reliable discharge planning.

A discharge plan must synthesize a patient's entire hospital course---diagnoses, medications, procedures, and follow-up needs---into a coherent, actionable packet. Errors or omissions in this document can lead to preventable readmissions and adverse drug events\cite{ref18}. Given this, Large Language Models (LLMs) can be a helpful, promising tool, which demonstrate remarkable proficiency in natural language understanding, summarization, and medical reasoning\cite{ref19,ref20}. However, their adoption in clinical practice faces the problem concerning safety and reliability: LLMs are fundamentally probabilistic engines prone to hallucination (fabricating facts)\cite{ref21,ref22,ref23} and miscalibration (expressing high confidence in incorrect outputs)\cite{ref24}. In a high-stake medical environment, a "confidently unsafe" AI assistant can be more dangerous than an incompetent one, as it invites automation bias with which clinicians may inadvertently trust flawed recommendations\cite{ref25,ref26}.

Traditional approaches to mitigating these risks have largely relied on "human-in-the-loop" verification, where a clinician reviews every AI output\cite{ref27}. While necessary for safety, this bottleneck caps the scalability of AI solutions\cite{ref28,ref29}. To truly unlock the value of clinical AI, we must move towards agentic AI---systems capable of autonomous reasoning, planning, and, critically, self-improving\cite{ref30}. It does not simply output the first token sequence that comes to mind (System 1 thinking). Instead, it engages in "System 2" thinking: it drafts a plan, observes its own output, evaluates it against external constraints, and refines it before ultimately presenting it to the user\cite{ref31,ref32}. This "thought-action-observation" loop mimics the cognitive process of a human clinician who reviews their own notes before signing them.



In this light, we present the Planner–Auditor framework, a dual-component architecture where the Planner agent drafts discharge plans utilizing FHIR-structured patient context from the MIMIC-IV-on-FHIR dataset and optional clinical guidelines. Complementing the generative Planner, the Auditor acts as a deterministic, rule-based observer that evaluates task coverage, calibration, and drift without directly modifying the plan. Safety and reliability are achieved by channeling Auditor signals into a two-tier control mechanism: a self-improvement loop for immediate regeneration within the request, and a discrepancy buffer that captures and replays high-confidence omissions for offline repair. The system is implemented via FastAPI endpoints and an ablation harness, supporting rigorous comparison across five configurations: baseline, context caching, self-improvement (SI), SI + context caching, and buffer replay. To summarize, our contributions are as follows:

\begin{enumerate}
\item FHIR-native discharge planning pipeline that ingests MIMIC-IV FHIR dataset, with deterministic snapshotting and LLM planning, evaluated using a retrospective simulation approach.
\item Quantified effect of enhanced configurations vs. baseline in ablations (self-improve drives the main coverage/calibration gains observed while context caching reduces latency while preserving performance).
\item Calibration gains: Reduction of high-confidence misses via self-improvement configuration.
\item Discrepancy buffer replay: target the hardest high-confidence/low-coverage episodes (residual failures).
\item FastAPI-based evaluation harness for systematic ablations across the five active configs (baseline, context caching, self-improvement, cache+SI, and optional buffer replay).
\end{enumerate}

\section*{2. Methods}

\subsection*{2.1 Dataset and Cohort Selection}



Our system was developed and evaluated using a retrospective simulation approach on the MIMIC-IV-on-FHIR dataset [33], encoded into FHIR R4 resources including Patient, Encounter, Condition, MedicationRequest, Observation, and Procedure fields. The evaluation focused on a retrospective cohort of 50 patients, with identifiers extracted sequentially from the dataset subject to configuration limits. To ensure sufficient clinical context for valid discharge planning, we applied specific inclusion criteria requiring that each patient’s clinical snapshot contain at least one active condition, medication, recent observation, procedure, or encounter.

\subsection*{2.2 System Architecture}

\begin{figure*}[t]
    \centering
    \setlength{\abovecaptionskip}{0pt}
    \includegraphics[width=\textwidth]{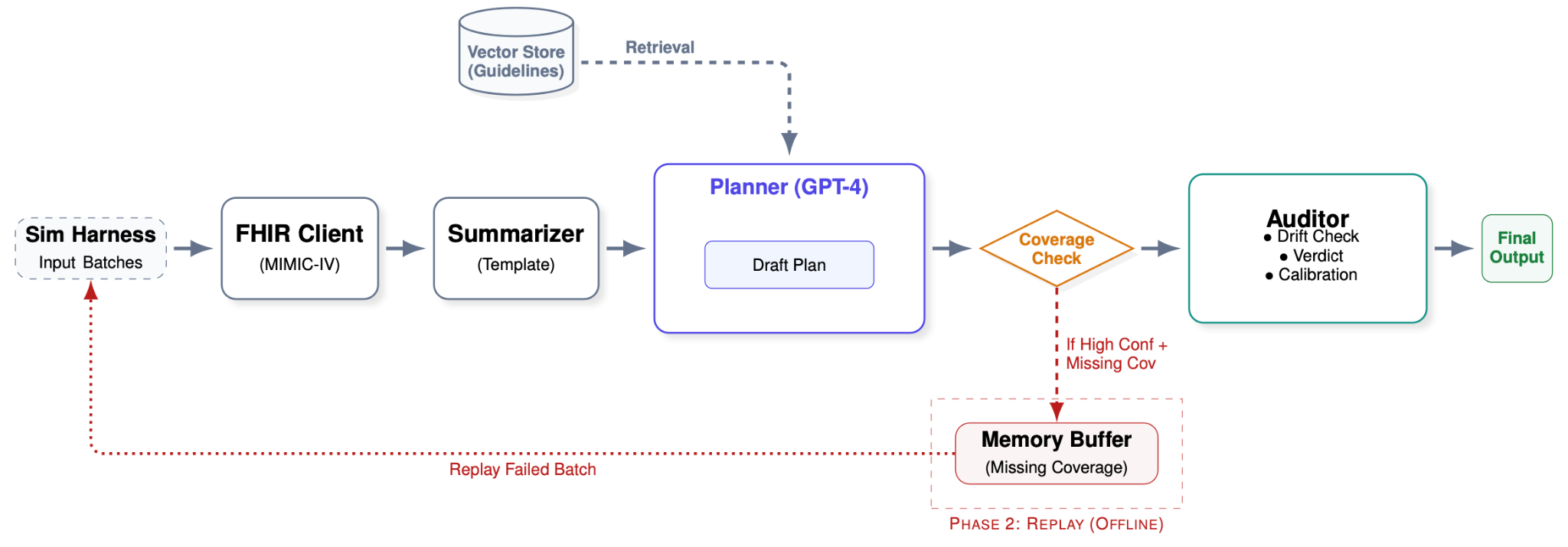} 
    \caption{Overview of Planner-Auditor architecture. The workflow begins with FHIR data retrieval and summarization via the simulation harness. The Planner Agent (GPT-4o-mini) generates a draft plan. A coverage check gate evaluates the draft; high‐confidence plans with missing coverage are flagged into a memory buffer. The downstream Auditing Agent scores coverage, calibration, and drift and records a verdict without blocking real-time generation. In the cross-episode replay step, buffered failures are re-injected for offline refinement to repair the flagged cases.}
    \label{fig:fig1_system_architecture}
\end{figure*}

Our discharge planning system is built on a retrieval-augmented generation (RAG) pipeline that interfaces with a FHIR R4 server to fetch patient data and generate structured discharge plans, as illustrated in Figure~\ref{fig:fig1_system_architecture}.

\subsubsection*{2.2.1 FHIR Data Access and Context Summarization}

Upon receiving a patient\_id, the system performs the following steps:

\begin{enumerate}
\item FHIR Data Fetch: The system uses a FHIRClient to query a HAPI FHIR server, retrieving a patient's active data bundle via FHIR R4 REST API. This includes Patient, Encounter, Condition, MedicationRequest, and Observation resources, all sourced from MIMIC-IV-on-FHIR data pre-loaded into the server.
\item Deterministic Summarization: We use a non-LLM, template-based summarizer to deterministically flatten FHIR resources into a text narrative. This process ensures the medication and diagnosis lists are an exact reflection of the record without hallucination. A SummaryGenerator processes raw FHIR bundles to create structured PatientSnapshot objects, which include: (1) a text summary, (2) structured JSON for active conditions, medications, observations, procedures, and encounters, and (3) metadata. This helps guarantee reproducibility and consistency in evaluation runs.
\item Guideline Retrieval: Upon preloading guideline chunks into the in‐memory vector store, the pipeline can query them and return top‐k snippets; otherwise it returns empty and runs on patient context alone.
\item Context Construction: The patient summary and retrieved guidelines are combined into a LLM's prompt context.
\end{enumerate}

\subsubsection*{2.2.2 The Planner Agent}

The discharge planning system uses a GPT-4o-mini Planner to generate a structured ActionPlan JSON object, which covers four categories: Follow-up Appointments, Medication Reconciliation, Patient Education, and Symptom Monitoring. The core logic is a RAGPipeline that integrates all contexts constructed above with an optional self-improvement loop that regenerates when necessary. Each request produces a JSON plan where every action includes a type, details, and deadline\_hours. The system supports multiple LLM backends (OpenAI, Anthropic, or HuggingFace).

\subsubsection*{2.2.3 The Auditing Agent}

The auditing agent is a deterministic Python module that inspects the JSON output from the Planner. It performs three specific validations:

1. Coverage Assessment:

It iterates through the list of generated actions and checks for the presence of all four mandatory tags using a CoverageChecker logic (if any category is missing, it flags a Coverage Violation):

\begin{equation*}
\begin{aligned}
\text{Coverage}_{\text{All}}
&= \mathbb{I}\Big(
   \text{Has}_{\text{FollowUp}}
   \wedge \text{Has}_{\text{Meds}}
   \wedge \text{Has}_{\text{Education}} \\
&\qquad\wedge \text{Has}_{\text{Monitoring}}
\Big)
\end{aligned}
\end{equation*}

2. Drift:

To ensure the model is not collapsing into a single output mode, it tracks the distribution of action types across the dataset. It computes L1 drift on action-type distribution. If L1 > 0.4 (handpicked by us), a drift warning is logged. It then outputs a verdict (PASS or FAIL) and a set of specific error messages (e.g., "Plan is missing Patient Education").

3. Calibration: To tracks Brier/ECE and overconfident errors; logs a PASS/FAIL verdict and violations but does not block generation.

\subsection*{2.3 The Two-Tier Self-Improvement Mechanism}

Self-improving loop consists of a lightweight internal refinement plus a cross-episode buffer:

\begin{enumerate}
\item Tier 1 (Within-episode refinement): When enable\_self\_improve is on, the RAG pipeline may regenerate a plan within the same request based on its own internal checks. The prompt is re-run with the patient context and prior draft to produce an improved plan; passes are recorded in RAG results.
\item Tier 2 (Discrepancy Buffer): If a plan is high-confidence ($\geq$0.8) but fails coverage (coverage\_gate=False), the case is logged to a buffer (buffer\_flag=True). Those "hard" cases can be replayed in a separate buffer\_replay run for focused retry.
\end{enumerate}

\subsection*{2.4 Evaluation Metrics and Ablation Configurations}

We run discrete-event simulations over MIMIC-IV-on-FHIR patients. For each patient, the system (1) fetches FHIR bundles, (2) generates a patient snapshot, (3) run the RAG planner (with/without self-improve, with/without cache), (4) assesses the coverage/calibration/drift of the plan by Auditor, and (5) log per-sample JSONs. Empty snapshots are skipped. Current configs include the following:

\begin{enumerate}
\item Baseline: one-shot, no self-improve, no context cache.
\item Context Cache: caching on, no self-improve.
\item Self-Improve: self-improve on, no cache.
\item Cache + Self-Improve: Both on.
\item Buffer Replay: reruns buffered high-conf/low-coverage cases (if invoked).
\end{enumerate}

Primary Metrics:

\begin{enumerate}
\item Multi-Task Coverage: Percentage of plans containing all four required categories.
\item Brier Score: Calibration of reported confidence vs. correctness proxy (coverage\_all), where p\textsubscript{i} is the predicted confidence and y\textsubscript{i} is the observed binary outcome (1 if correct, 0 if incorrect; the lower is better).

\begin{equation*}
\text{Brier}=\frac{1}{N}\sum_{i=1}^{N}\left(p_i-y_i\right)^{2}
\end{equation*}

\item Expected Calibration Error (ECE): Weighted average of the difference between confidence and accuracy across bins.

\begin{equation*}
\text{ECE}=\sum_{b=1}^{B}\frac{n_b}{N}\left|\text{acc}_b-\text{conf}_b\right|
\end{equation*}

or more detailed,

\begin{equation*}
\text{ECE}=\sum_{b=1}^{B}\frac{n_b}{N}\left|\frac{1}{n_b}\sum_{i\in B_b}y_i-\frac{b-0.5}{B}\right|
\end{equation*}

where B=10 is the number of bins, n\textsubscript{b} is the number of predictions in bin b, N is the total number of predictions, acc\textsubscript{b} is the observed accuracy in bin b, and conf\textsubscript{b} is the bin center (midpoint of the confidence interval).

\item High-Confidence Error Rate: proportion with confidence $\geq$0.8 but coverage\_all=False.
\item Latency/throughput: mean wall-clock per episode.
\item Drift (L1): action-type distribution shift.
\end{enumerate}

\section*{3. Results}

\subsection*{3.1 Plan Completeness and Multi-Task Coverage}

All enhanced configurations outperformed the baseline in discharge plan completeness, achieving substantially higher coverage of all required tasks. The baseline often produced incomplete plans -- only about 32\% of its discharge summaries contained all four essential tasks, underscoring frequent omissions in follow-up, medication, education, or monitoring content (Figure 2, Table 1). In contrast, every improved method yielded far more comprehensive plans. For instance, incorporating contextual memory and self-improvement enabled 52\% and 86\%, respectively, of episodes to include \textit{all} four tasks, effectively doubling the full coverage rate relative to baseline. Excluding the buffer replay, the best-performing strategy, combining context caching with self-improvement, attained highest multi-task coverage and lowest ECE (0.034) while maintaining efficiency.

Not only did overall plan completeness improve, but each individual task category was addressed more consistently by these approaches. As shown in Table 1 and Figure 2, the baseline particularly struggled with patient education and monitoring recommendations -- these appeared in far fewer than half of baseline plans. Methods with self-improvement corrected this deficit: most of them added the missing education and arranged appropriate monitoring when the initial plan lacked them. Follow-up appointments and medication prescriptions, which the baseline included more often, still saw gains with the improved configurations, reaching completion rates close to 100\%. These results demonstrate that our interventions (context caching and iterative self-refinement) led to more complete and thorough discharge plans, covering the full spectrum of post-discharge tasks at rates dramatically higher than the baseline. The combination of context caching with self-improvement was especially effective, suggesting their benefits are complementary -- context reuse provides relevant prior information, and self-improvement ensures no task is overlooked. Together, they produce discharge instructions that excel in addressing patient needs.

\begin{table*}[hbt!]
\setlength{\abovecaptionskip}{5pt}
\caption{Ablation summary (completion, coverage, calibration, latency). The baseline is weak in education and monitoring (54\%), yielding low coverage (32\%). Context caching improves completeness (52\% coverage, mainly monitoring/medication) and lowers latency (17.42s $\rightarrow$ 11.83s). Self-improvement reaches high coverage (86\%) but increases latency (19.65s). Combining cache + self-improvement maintains high coverage (86\%), improves latency (18.70s), and achieves the best ECE (0.034), suggesting a positive synergy. Buffer replay achieves perfect coverage (100\%) but has the highest latency (27.78s) and a worse ECE (0.107).}
\label{tab:table1}
\centering
\resizebox{13cm}{!}{%
\begin{tabular}{lcccccccccc}
\toprule
& & \multicolumn{5}{c}{\textbf{Success \& Coverage Rates}} & \multicolumn{2}{c}{\textbf{Calibration}} & \multicolumn{2}{c}{\textbf{Efficiency}} \\
\cline{3-7}\cline{8-9}\cline{10-11}
\textbf{Configuration} & $N$ & \textbf{Cover.} & \textbf{F/Up} & \textbf{Meds} & \textbf{Edu.} & \textbf{Monitor} & \textbf{Brier} & \textbf{ECE} & \textbf{Lat. (s)} & \textbf{Ep./min} \\
\midrule
Baseline & 50 & 0.320 & 0.960 & 0.800 & 0.540 & 0.540 & 0.544 & 0.564 & 17.42 & 3.16 \\
Context Cache & 50 & 0.520 & 0.960 & 0.900 & 0.580 & 0.720 & 0.382 & 0.356 & 11.83 & 4.42 \\
Self Improve & 50 & 0.860 & 1.000 & 0.980 & 0.940 & 0.880 & 0.126 & 0.062 & 19.65 & 2.77 \\
Cache + Self Imp. & 50 & 0.860 & 1.000 & 0.960 & 0.980 & 0.920 & 0.123 & 0.034 & 18.70 & 2.95 \\
Buffer Replay & 7 & 1.000 & 1.000 & 1.000 & 1.000 & 1.000 & 0.017 & 0.107 & 27.78 & 2.09 \\
\bottomrule
\end{tabular}%
}
\end{table*}

\begin{figure*}[t]
    \centering
    \includegraphics[width=\textwidth,height=0.32\textheight,keepaspectratio]{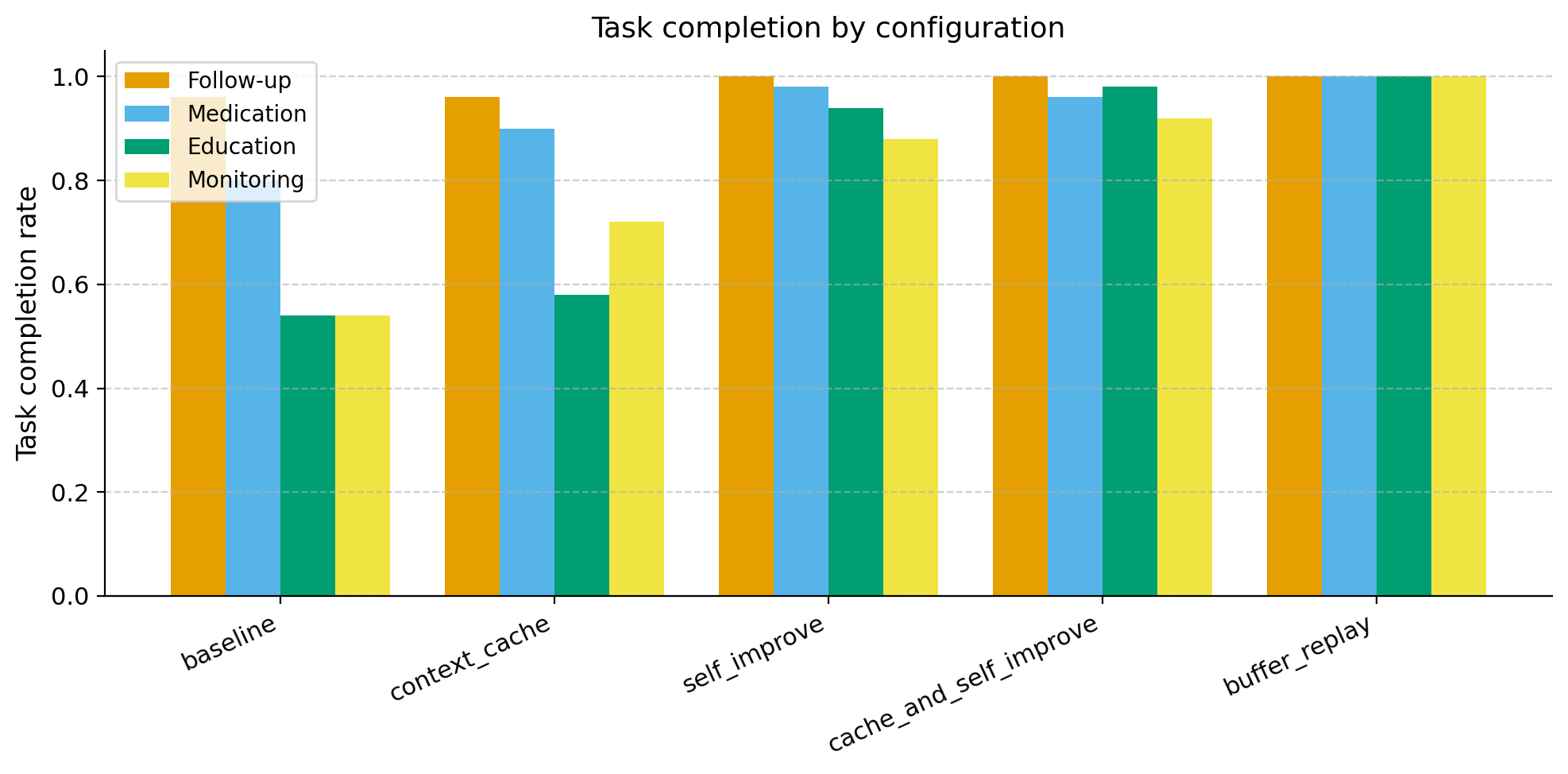}
    \par\smallskip
    {\small (a)}
    \par\medskip
    \includegraphics[width=\textwidth,height=0.32\textheight,keepaspectratio]{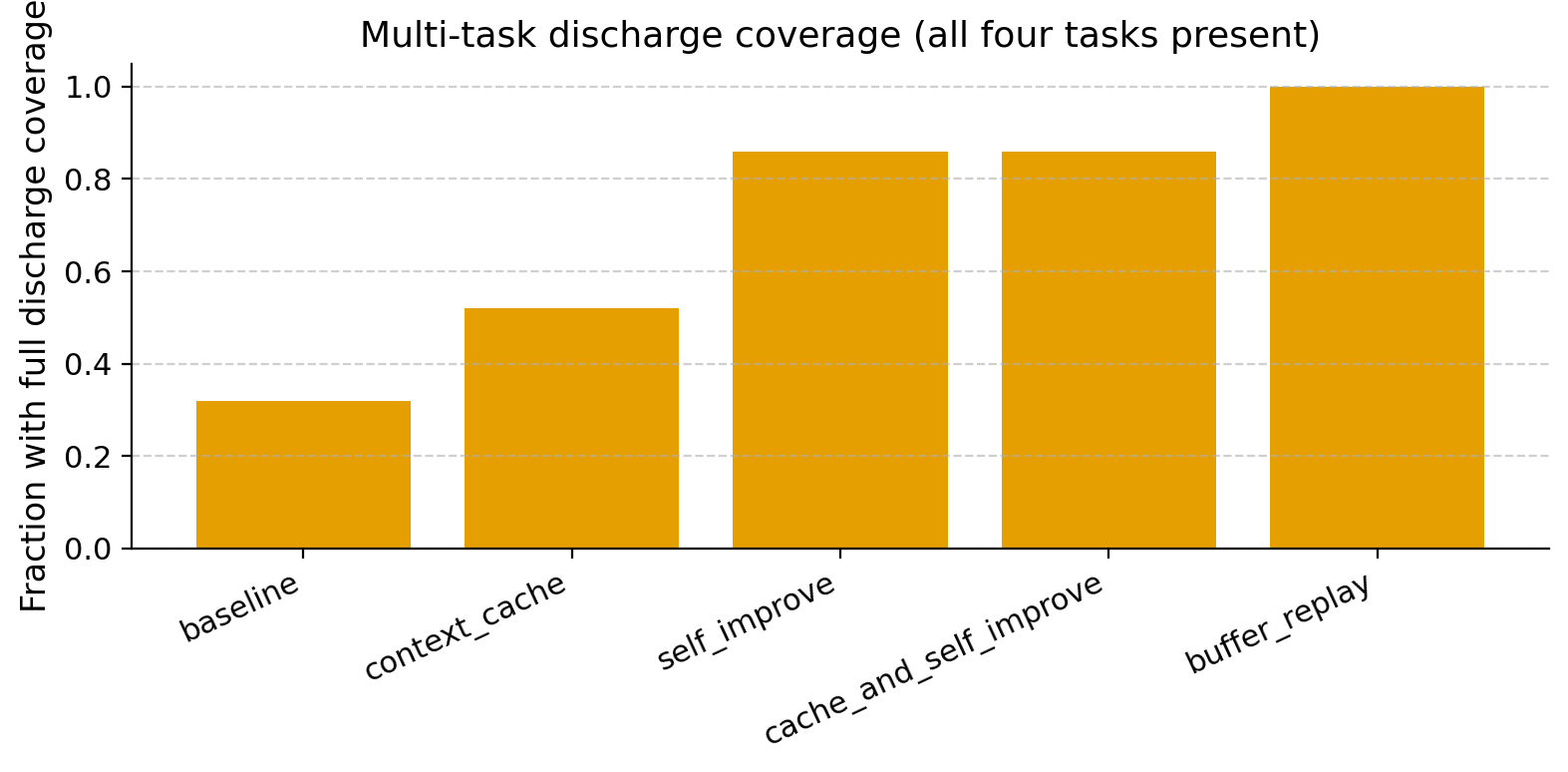}
    \par\smallskip
    {\small (b)}
    \caption{(a) Per-episode completion rates for four discharge planning tasks (follow-up, medication, education, monitoring) across system configurations. Follow-up is near ceiling (96-100\%). Baseline fails most often on education/monitoring (both 54\%). Caching moderately improves them, while self-improvement delivers the largest gains on them (education 54\% to 94\%, monitoring 54\% to 88\%). Cache + SI further boosts education (98\%) and monitoring (92\%). (b) Full discharge coverage (all 4 tasks present simultaneously) shows a staircase: Baseline 32\% to Cache 52\% to SI / Cache+SI 86\% to Buffer Replay 100\%. Caching provides a meaningful quality gain and improves latency. Cache + SI matches SI's coverage (86\%) more efficiently. Buffer Replay achieves 100\% coverage (cautious interpretation due to small N=7 and high latency cost).}
    \label{fig:fig2}
\end{figure*}

\subsection*{3.2 Safety and Calibration}

Adding context caching and self-improving loop not only increased completeness but also safety by reducing high-confidence omissions and better aligning the model's confidence with its performance. The baseline frequently exhibited \textit{high-confidence errors} -- cases where it was confident in its discharge plan despite missing important tasks. According to the auditor checks, in roughly 66\% of episodes, the baseline model failed to include one or more required tasks while still expressing high confidence in its output. This scenario is risky in a clinical context, as the model might wrongly assure providers that the plan is sufficient when it is in fact not. All our improved configurations dramatically lowered this high-confidence error rate, as displayed in Figure 3a. Notably, the self-improving based methods eliminated most confident omissions while buffer replay closed the remaining gap. Even the simpler context caching alone decently reduced high-confidence error rate from 66\% to 46\% relative to baseline, indicating that supplying memory from prior cases helped the model avoid certain mistakes or overconfidence.

In addition to reducing errors, the variant configurations also yielded better calibration of the model's confidence. According to Table 1 and Figure 3b, the baseline's Brier score was the highest (0.544), and its ECE (expected calibration error) was worst among all configurations (0.564). By contrast, all enhanced configurations show markedly lower Brier scores and ECE values. The self-improve-based methods achieved the lowest calibration error, indicating that the model's self-reported confidence more faithfully reflected the completeness of its plan. In practical terms, a clinician can have greater trust that when the improved system claims a discharge plan is complete, it genuinely \textit{is} complete. Overall, our results demonstrate that iterative self-refinement and context awareness not only produce more complete plans, but also make the outputs safer -- the model is far less likely to "fail silently" by omitting content it falsely believes it included.

\begin{table*}[hbt!]
\setlength{\abovecaptionskip}{5pt}
\caption{Safety metrics (high-confidence errors, auditor failures, violations). Safety improvements are large and systematic: auditor failure rate falls from 66\% (baseline) to 46\% (context\_cache) to 14\% (self-improve variants) and 0\% (buffer\_replay). High confidence errors and coverage violations also drop consistently. The "failure rate" counts every auditor fail; the "high‐confidence error rate" counts only the fails with confidence above the threshold. In this case, the two appeared to coincide.}
\label{tab:table2}
\centering
\begin{threeparttable}
\resizebox{13cm}{!}{%
\begin{tabular}{lcccccccc}
\toprule
& & \multicolumn{3}{c}{\textbf{Planner Outcomes}} & \multicolumn{2}{c}{\textbf{Violation Types}} & \multicolumn{2}{c}{\textbf{Critical Safety}} \\
\cline{3-5}\cline{6-7}\cline{8-9}
\textbf{Configuration} & $N$ & \textbf{Pass} & \textbf{Fail} & \textbf{Fail Rate} & \textbf{Coverage} & \textbf{Drift ($L_1$)} & \textbf{HC Err}\tnote{a} & \textbf{Avg. Conf.} \\
\midrule
Baseline & 50 & 17 & 33 & 0.660 & 33 & 0.000 & 33 & 0.881 \\
Context Cache & 50 & 27 & 23 & 0.460 & 23 & 0.058 & 23 & 0.872 \\
Self Improve & 50 & 43 & 7 & 0.140 & 7 & 0.125 & 7 & 0.864 \\
Cache + Self Imp. & 50 & 43 & 7 & 0.140 & 7 & 0.103 & 7 & 0.875 \\
Buffer Replay & 7 & 7 & 0 & 0.000 & 0 & 0.075 & 0 & 0.871 \\
\bottomrule
\end{tabular}%
}
\begin{tablenotes}[flushleft]
\item[a] High Confidence Errors: Instances where the agent was confident but incorrect.
\end{tablenotes}
\end{threeparttable}
\end{table*}

\begin{figure*}[t]
    \centering
    \includegraphics[width=\textwidth,height=0.30\textheight,keepaspectratio]{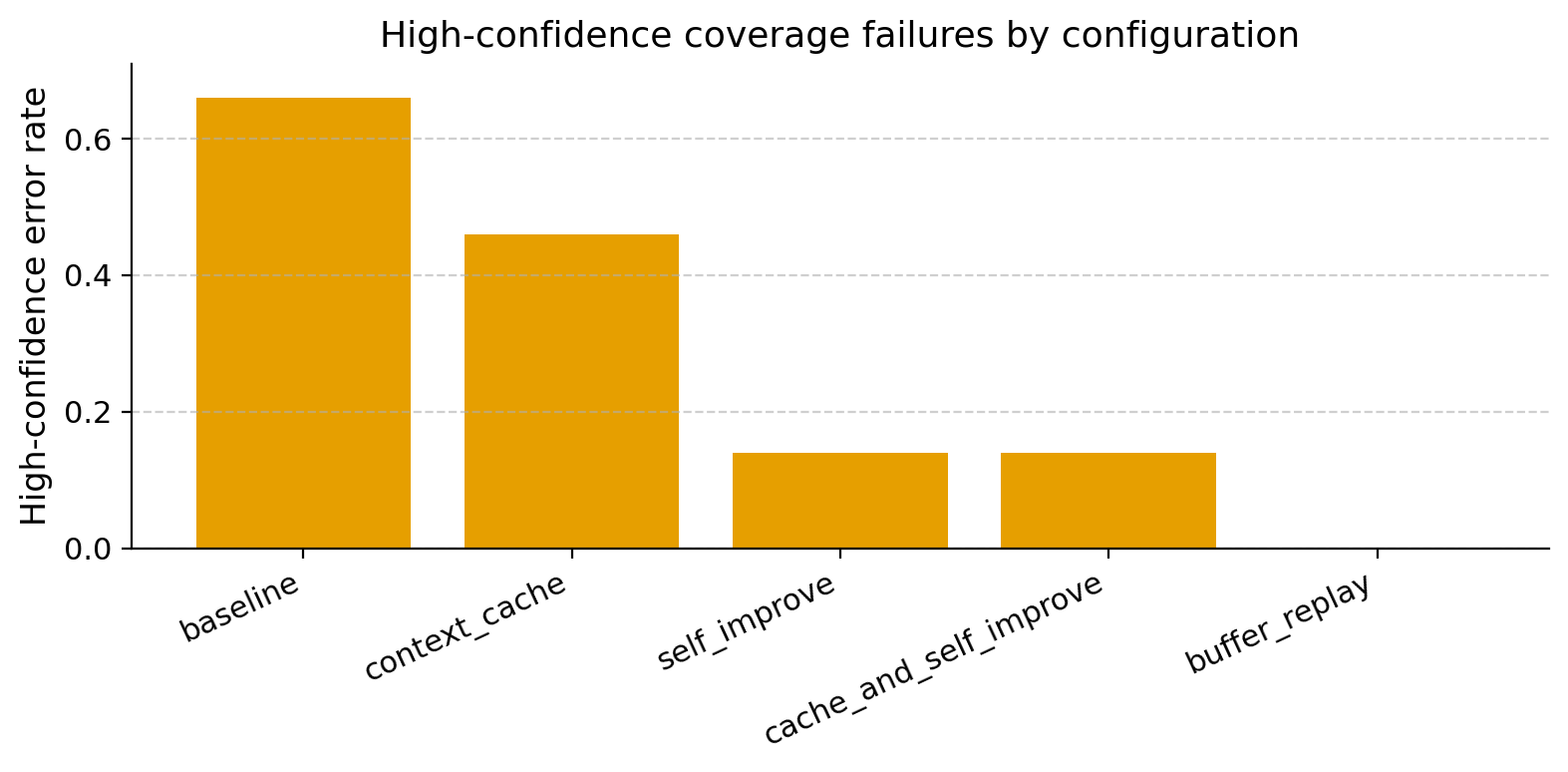}
    \par\smallskip
    {\small (a)}
    \par\medskip
    \includegraphics[width=\textwidth,height=0.30\textheight,keepaspectratio]{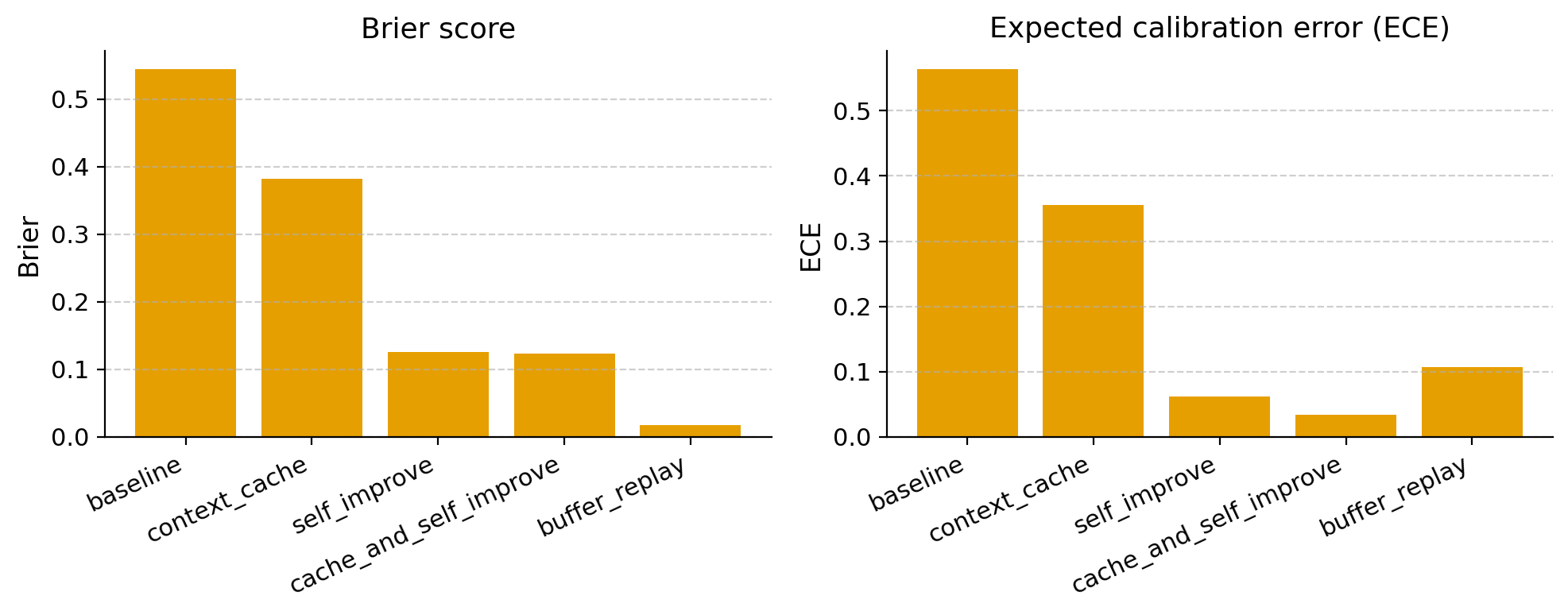}
    \par\smallskip
    {\small (b)}
    \caption{(a) High-confidence coverage failures by configuration. High-confidence error rate (fraction of episodes with coverage failure assigned high confidence) drops with completeness mechanisms: baseline (66\%) $\rightarrow$ context caching (46\%) $\rightarrow$ self-improvement (14\%) and cache+SI (14\%), with buffer replay being 0\%. (b) Calibration quality. Left, Brier score; right, ECE. Calibration improves sharply with completeness mechanisms: baseline is worst (Brier 0.544, ECE 0.564), improving with context caching (Brier 0.382, ECE 0.356) and substantially with self-improvement (Brier 0.126, ECE 0.062). Cache + self-improve achieves the best ECE (0.034). Buffer replay has the lowest Brier (0.017) but worse ECE (0.107).}
    \label{fig:fig3}
\end{figure*}

\subsection*{3.3 Efficiency Trade-offs}

On the other hand, the benefits of improved completeness and safety come with efficiency trade-offs. We quantified computational efficiency using mean end-to-end latency per episode and throughput (episodes per minute), and visualized the quality--latency and calibration-latency trade-off in Figure 4a and 4b, respectively. Context caching is both faster and higher-throughput than the baseline in this ablation run. Specifically, mean latency decreases from 17.42 s (baseline) to 11.83 s (context\_cache), a reduction of 5.59 s, while throughput increases from 3.16 to 4.42 episodes/min in Table 1. Importantly, this speedup occurs alongside a substantial quality improvement: full coverage rises from 32\% to 52\%, suggesting that the baseline's inefficiency may reflect pipeline overhead or longer/less-direct generations that are mitigated when relevant context is reused.

Moving from caching to iterative refinement introduces a more conventional cost--benefit trade-off. Self-improvement increases full coverage to 86\%, but with higher mean latency (19.65 s) and reduced throughput (2.77 episodes/min) relative to context\_cache. Combining caching with self-improvement partially recovers efficiency without sacrificing completeness: cache\_and\_self\_improve maintains the same full coverage (86\%) while reducing latency to 18.70 s (an absolute decrease of 0.95 s compared to self\_improve) and increasing throughput to 2.95 episodes/min in Figure 4a and 4b. This pattern indicates that caching can meaningfully reduce the overhead of refinement loops even when the quality target is unchanged.

Buffer replay represents the high-coverage extreme in this run (full coverage 100\%) but is also the most expensive configuration, with mean latency 27.78 s and throughput 2.09 episodes/min according to Figure 4a and 4b. Because buffer\_replay is only evaluated on N=7 episodes, we treat it as an exploratory point: it is consistent with a regime that can maximize completeness, but its efficiency---and even its stability---requires confirmation at larger scale before drawing deployment-level conclusions.

These trade-offs are summarized by the Pareto view in Figure 4c. The nondominated set (coverage $\uparrow$, latency $\downarrow$) consists of context\_cache, cache+self\_improve, and buffer\_replay---while baseline is dominated by context\_cache and self\_improve is dominated by cache+SI (less latency with no worse/even better performance). Taken together, the efficiency analysis suggests a practical operating continuum: context\_cache offers a strong "default" setting with simultaneous speed and quality gains over baseline, cache\_and\_self\_improve offers high completeness with improved efficiency over self\_improve, and buffer\_replay is a high-completeness, high-latency regime that warrants further scaling and robustness checks.

\begin{figure*}[t]
    \centering
    \begin{minipage}{0.49\textwidth}
        \centering
        \includegraphics[width=\linewidth,height=0.28\textheight,keepaspectratio]{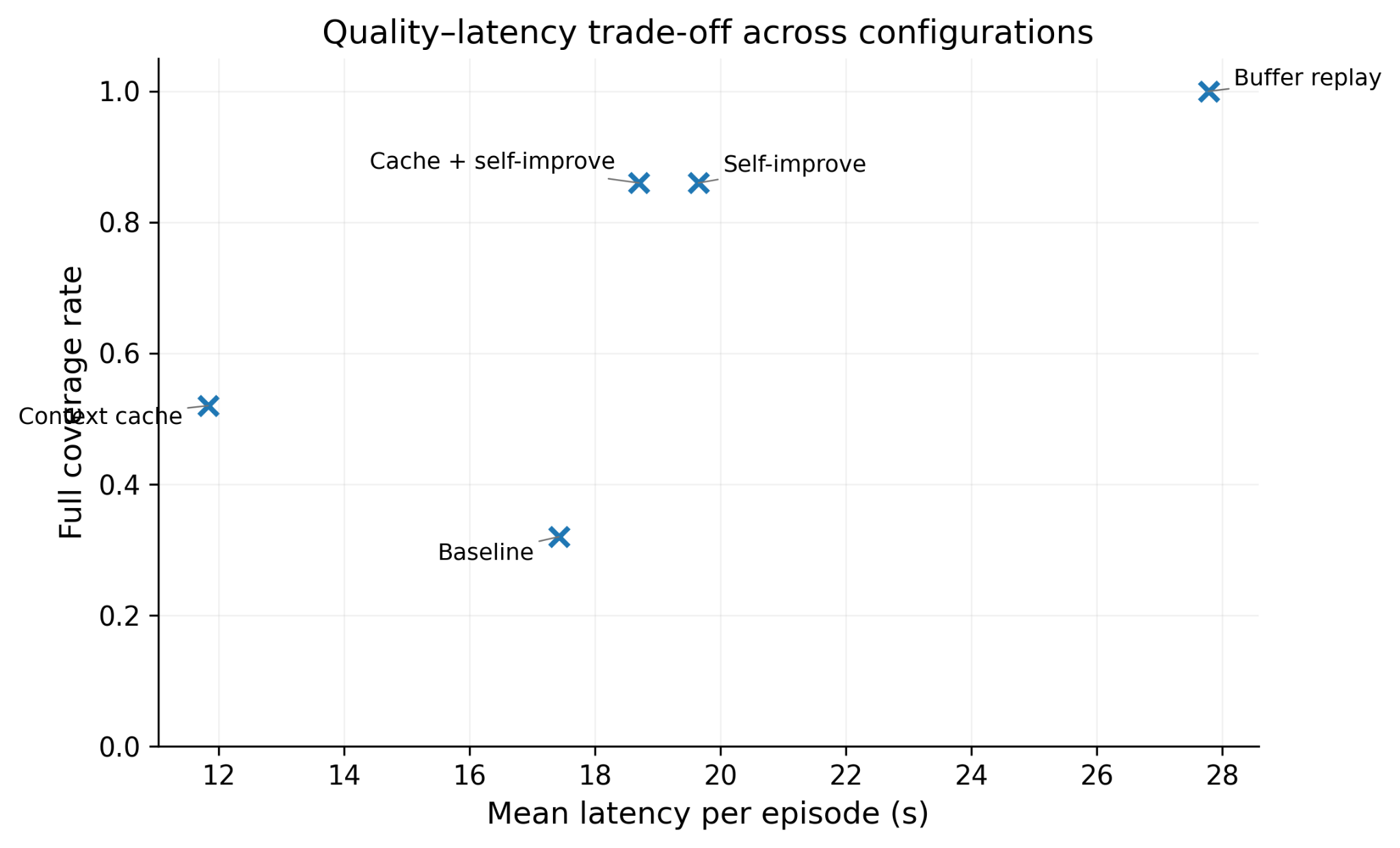}
        \par\smallskip
        {\small (a)}
    \end{minipage}
    \hfill
    \begin{minipage}{0.49\textwidth}
        \centering
        \includegraphics[width=\linewidth,height=0.28\textheight,keepaspectratio]{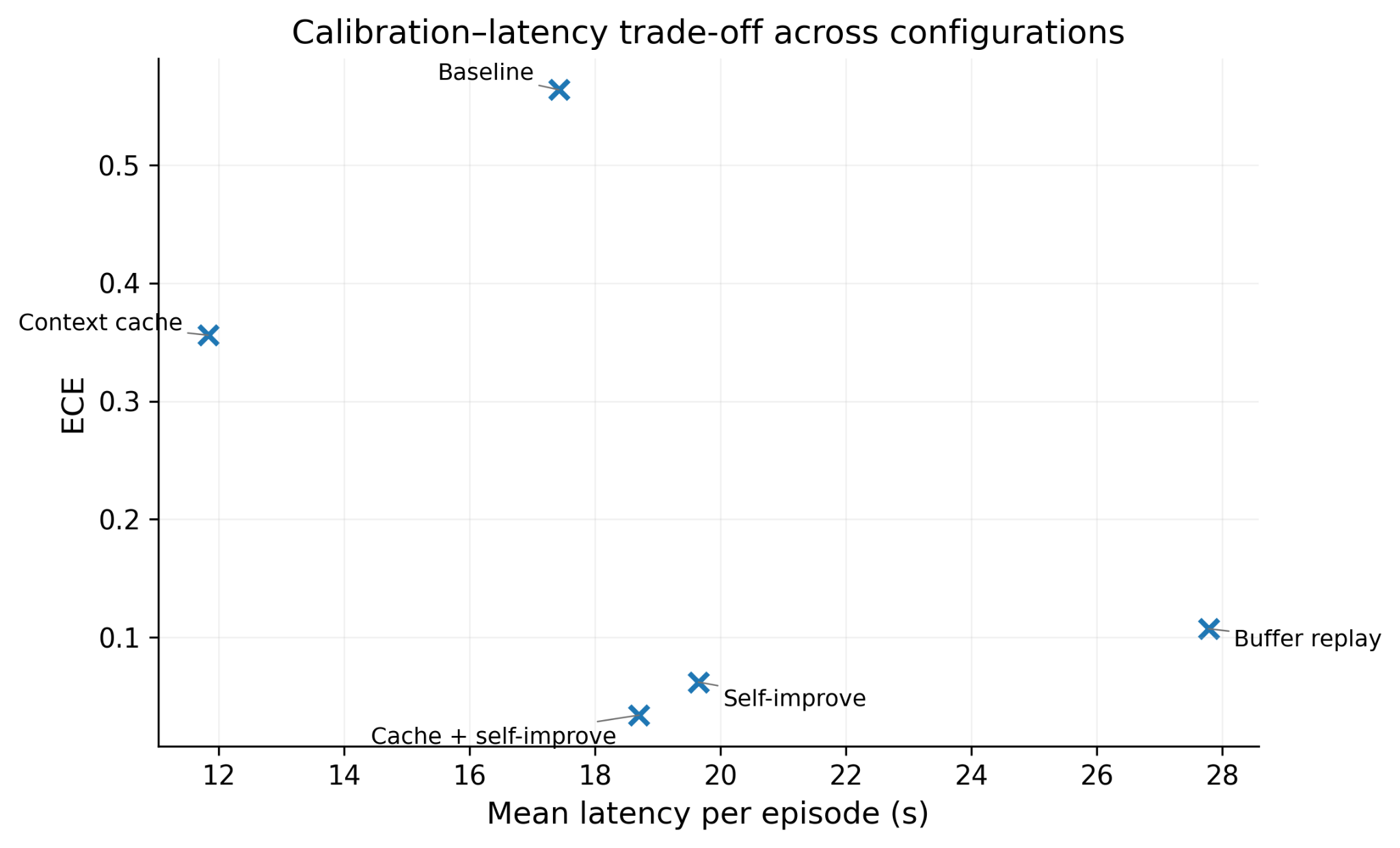}
        \par\smallskip
        {\small (b)}
    \end{minipage}
    \par\medskip
    \includegraphics[width=0.70\textwidth,height=0.33\textheight,keepaspectratio]{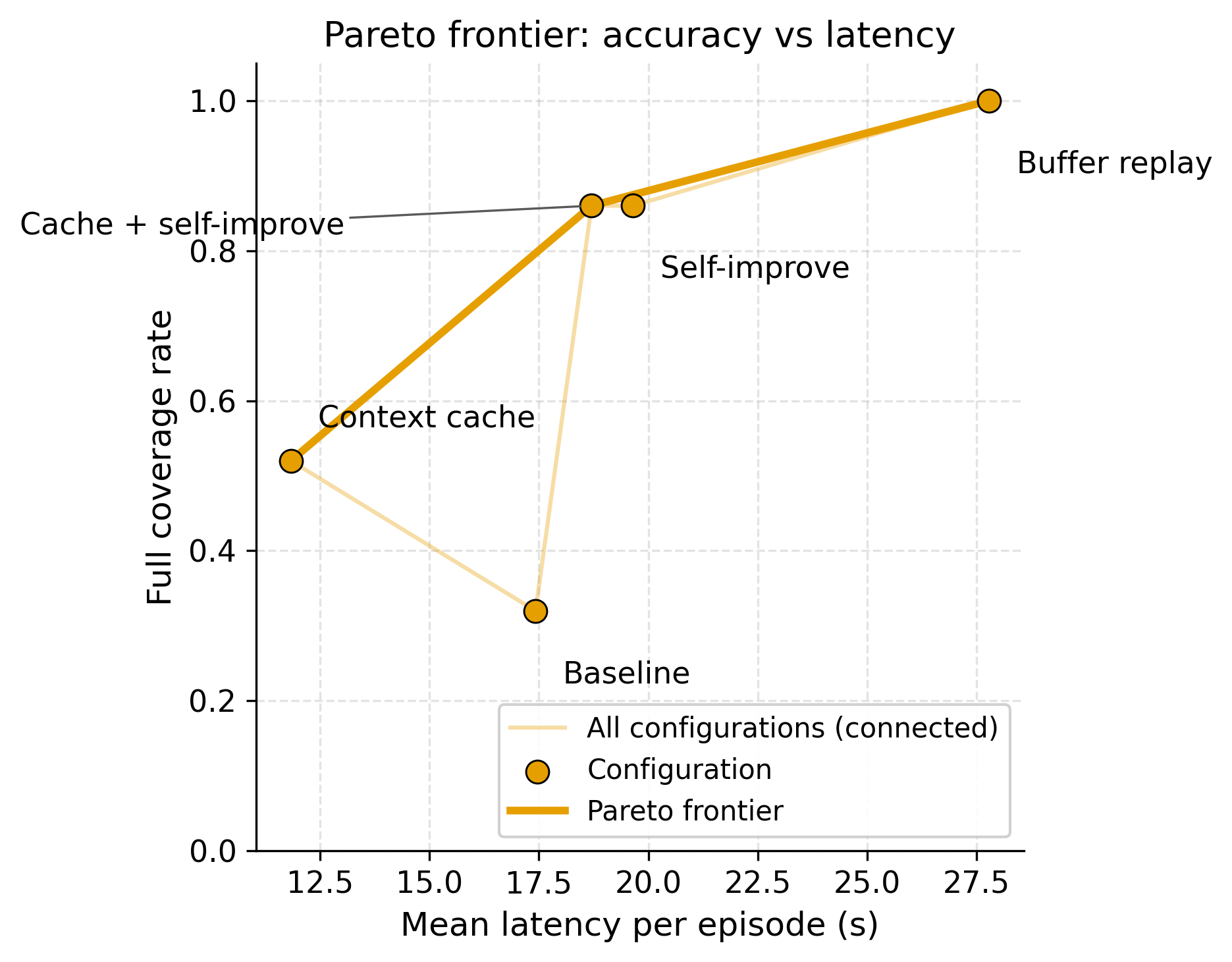}
    \par\smallskip
    {\small (c)}
    \caption{(a) Quality--latency trade-off. Context cache dominates baseline. Self-improvement significantly boosts quality (86\% coverage) with latency overhead (19.65s). Cache + SI improves SI: same coverage (86\%) at lower latency (18.70s). Buffer replay achieves perfect coverage (100\%) but with highest latency (27.78s). (b) Calibration--latency trade-off. ECE decreases dramatically with SI (0.564 $\rightarrow$ 0.062) and Cache+SI (0.034). Buffer replay's ECE (0.107) is worse than SI variants with even higher latency. Cache+SI is the best-calibrated high-coverage operating point. (c) Pareto frontier: accuracy vs latency. Line in dark orange marks the Pareto-efficient frontier, representing configurations minimizing latency while maximizing full coverage.}
    \label{fig:fig4}
\end{figure*}

\begin{figure*}[t]
    \centering
    \includegraphics[width=\textwidth,height=0.55\textheight,keepaspectratio]{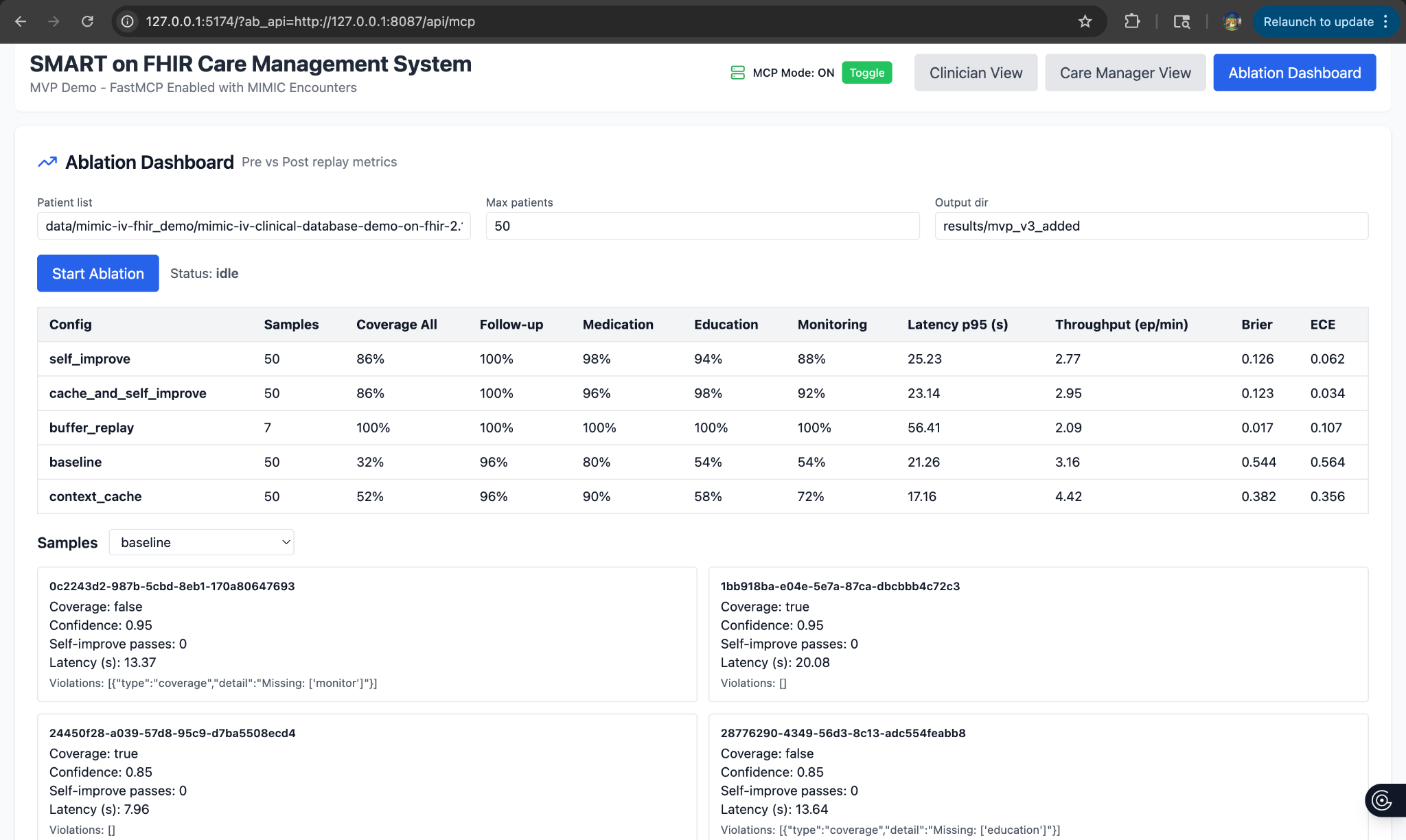}
    \caption{Front-end evaluation dashboard for discharge planning system built by React and fastMCP. Interactive web interface shows comparative metrics (coverage, latency, calibration), per-sample results, violations, and real-time execution logs across multiple system configurations (baseline, context cache, self-improvement, and combinations). The dashboard connects to a FastAPI evaluation backend.}
    \label{fig:fig5}
\end{figure*}


\section*{4. Discussion}

Wrapping the planner with deterministic auditing and buffer replay materially improves robustness without retraining. It also reduces dangerous omissions in discharge plans and yields better confidence alignment with observed correctness, lowering the risk of unnoticed errors. These gains come with higher runtime, so there is a quality--latency trade-off.

\subsection*{4.1 The "Auditor + Buffer" Safety Net}

In this light, the combination of the auditor and discrepancy buffer creates a robust deployment strategy. In a real-world hospital, this architecture allows for a "triage" of AI outputs as safety comes from feeding its signals back into generation and replaying high‐confidence failures:

\begin{enumerate}
\item Green Lane: Auditor Pass $\rightarrow$ surface to clinician.
\item Yellow Lane: Auditor FAIL + low confidence $\rightarrow$ route to manual review.
\item Red: Auditor FAIL + high confidence $\rightarrow$ log into the discrepancy buffer and replay. These are the dangerous errors that our system intends to catch before they reach users.
\end{enumerate}

In our runs, buffer replay resolved the remaining high‐confidence/low‐coverage cases, showing that many "hard" errors are brittle and can be fixed with deliberate retry.

\subsection*{4.2 Clinical and Operational Implications}

Self‐improvement variants run slower than baseline due to the extra refinement pass, but discharge planning is asynchronous, so the latency increase may be acceptable in practice. The main benefit is fewer high‐confidence coverage misses and better calibration, moving the system closer to a usable clinical assistant rather than an unsafe prototype.

\subsection*{4.3 Limitations}

First, we used a specific simulated discharge scenario and a retrospective cohort (MIMIC-IV-FHIR); validation on real-world clinical cases would be essential to assess impact on clinician workflow, with adding (for example) Clinical Research Datamart (CRDM) datasets from Duke University School of Medicine as a next step. Second, our evaluation focused on the presence of four predefined task types in discharge plans, not directly measuring quality, clarity, or personalization. Third, our auditor flagged dangerous complacency based on task omission and confidence, potentially missing other safety issues such as incorrect medical content. Future work should incorporate, if possible, clinical accuracy checks and human clinician feedback. Finally, the sample size (N\textasciitilde{}50) we used is sufficient for a demo but must be scaled for definitive statistical power.

\section*{5. Conclusion}

This study highlights a promising direction for safer and more reliable AI-assisted clinical workflows. By using LLM agents' capabilities in iterative reflection and leveraging prior context to fill gaps, we showed these yield remarkable improvements in completeness and confidence calibration for discharge planning. With consideration of efficiency trade-offs and further refinement, such an approach could increase the reliability of AI assistants in healthcare, helping to ensure that no critical instructions are missed or wrongly executed in the transition from hospital to home.



\section*{ACKNOWLEDGEMENTS}
This project is supported in part by SPARK-ICU R01 grant R01GM139967 (Rishikesan Kamaleswaran).

\section*{AUTHOR CONTRIBUTIONS}
R.K and K.W. conceived the main framework. K.W. conducted the experimental studies, wrote the manuscript, and A.D reviewed and polished it. All authors contributed to the article and approved the submission of this manuscript.

\section*{CONFLICT OF INTEREST}
The authors declare no competing interests.

\section*{CODE AVAILABILITY}
The MIMIC-IV-FHIR dataset used in the experiment can be accessed upon request via PhysioNet. The source code for replicating the experiments will be made available on GitHub repository once the manuscript got accepted and published online.

\section*{Use of generative AI/LLM tools}
During the refinement stage of this manuscript, the authors used \emph{OpenAI's GPT5} solely to assist with language editing and polishing of wording and grammar.
The tool was not used to generate new scientific content, analyses, results, or conclusions.
All authors reviewed and edited the manuscript for the content.


\end{document}